\documentclass[runningheads]{llncs}
\usepackage[T1]{fontenc}
\usepackage{graphicx}
\usepackage{booktabs}
\usepackage{fancyvrb}
\usepackage{subfig}
\usepackage{pgfplots}
\PassOptionsToPackage{hyphens}{url}\usepackage{hyperref}
\definecolor{LimeGreen}{rgb}{0.2, 0.8, 0.2}
\definecolor{ProcessBlue}{rgb}{0.0, 0.72, 0.92}

\usepackage{amsmath}

\begin{document}
\title{HITgram: A Platform for Experimenting with $n$-gram Language Models}
\titlerunning{HITgram: A Platform for Experimenting $n$-gram LMs}
%

\author{Shibaranjani Dasgupta\inst{1} \and
Chandan Maity\inst{1} \and
Somdip Mukherjee\inst{1} \and
Rohan Singh\inst{1} \and
Diptendu Dutta\inst{2} \and 
Debasish Jana\inst{1}}
\authorrunning{S. Dasgupta et al.}
%
\institute{
  Heritage Institute Of Technology, Kolkata\\
  \email{\{shibaranjani.dasgupta.cse25, chandan.maity.cse25, \\ somdip.mukherjee.cse25, rohan.singh.cse25\}@heritageit.edu.in}\\
  \email{debasish.jana@heritageit.edu}
  \and 
  Aunwesha Knowledge Technologies Private Limited, Kolkata\\
  \email{dutta.diptendu@gmail.com}
}
\maketitle              
\begin{abstract}
Large language models (LLMs) are powerful but resource-intensive, limiting accessibility. HITgram addresses this gap by offering a lightweight platform for $n$-gram model experimentation, ideal for resource-constrained environments. It supports unigrams to 4-grams and incorporates features like context-sensitive weighting, Laplace smoothing, and dynamic corpus management to enhance prediction accuracy, even for unseen word sequences. Experiments demonstrate HITgram's efficiency, achieving 50,000 tokens/second and generating 2-grams from a 320MB corpus in 62 seconds. HITgram scales efficiently, constructing 4-grams from a 1GB file in under 298 seconds on an 8 GB RAM system. Planned enhancements include multilingual support, advanced smoothing, parallel processing, and model saving, further broadening its utility. 
\keywords{$n$-Gram Model \and Text Generation \and Smoothing \and Computational Linguistics \and Natural Language Processing \and Generative AI}
\end{abstract}

\section{Introduction}
\label{sec:intro}
Large language models (LLMs) have revolutionized natural language processing (NLP) with attention-based architectures like transformers~\cite{vaswani2017attention}, excelling in tasks like next-word prediction and conversational agents ~\cite{chatgpt}. However, their high computational demands limit accessibility, prompting renewed interest in lightweight alternatives~\cite{brown2020LMFewShotLearners}\cite{kaplan2020scaling} such as small language models (SLMs) and traditional $n$-gram models~\cite{jurafsky2024speech}. 

\vspace{1mm} 
\noindent\textbf{$n$-gram Models.}
These models are among the earliest and most widely used methods in NLP and computational linguistics~\cite{manning20introduction}. By conditioning on the preceding $(n-1)$ words, $n$-gram models predict the $n^{\text{th}}$ word through statistical analysis of word frequencies and co-occurrences in large text corpora. Their statistical nature allows for computational efficiency and ease of implementation. This probabilistic approach has found applications in various domains, including text generation, spell-checking, language modeling, and speech recognition. 

\noindent Despite advancements in deep learning, $n$-gram models~\cite{hu2023advancing}\cite{katsafados2024machine}\cite{malagutti2024role} remain relevant due to their balance between performance and computational demands. Their ability to function with smaller unlabeled datasets and deliver interpretable results makes them a viable option for various practical NLP tasks, particularly where larger models are impractical due to resource constraints or data insufficiency. In this paper, we develop HITgram with this goal in mind, offering an accessible platform for users to interact with $n$-gram models and explore their functionality in a hands-on manner. HITgram's lightweight architecture enables its use in various practical applications, including predictive text, autocomplete features, search engine optimization, and speech-to-text preprocessing. These capabilities are particularly valuable in domains like education, accessibility tools, and mobile devices, where computational resources are often constrained.

\vspace{1mm} 
\noindent\textbf{Limitations of Traditional $n$-grams.}
Although $n$-gram models are efficient and interpretable, a key limitation they face is data sparsity~\cite{allison2006another}, particularly as the context length increases. This issue necessitates vast amounts of training data to capture meaningful contextual relationships. Additionally, traditional $n$-gram models are restricted to fixed-length context windows, constraining their capacity to model long-range dependencies and complex language structures.

\noindent Another common issue with $n$-gram models is handling unseen word sequences. Without proper smoothing techniques, these models struggle to generate accurate predictions for rare or previously unseen $n$-grams. To address this, the HITgram platform offers various smoothing options viz., Laplace and Good-Turing, to mitigate data sparsity and improve prediction accuracy. Furthermore, the platform allows users to experiment with different $n$-gram configurations (such as bigrams and trigrams) with varying context lengths, to better understand and optimize language modeling performance.

\vspace{1mm}
\noindent\textbf{Contributions.} HITgram facilitates $n$-gram model experimentation in resource-limited settings. It provides tools for building, testing, and refining $n$-gram models on user-provided corpora, balancing efficiency with accessibility while addressing traditional model challenges.

\vspace{1mm}
\noindent\textbf{Organization of the Paper.} 
The remainder of the paper is organized as follows. Section~\ref{sec:relwork} reviews related work, discussing  challenges with language models. Section~\ref{sec:motivation} outlines the motivation for this work. In Section~\ref{sec:overview}, we detail the HITgram platform, its architecture and implementation. Section~\ref{sec:expt} presents the experimental results and observations. Finally, Section~\ref{sec:concl} concludes the paper with suggestions for future work.

\section{Background and Related Work}
\label{sec:relwork}
$n$-gram models remain crucial in NLP, offering lightweight, computationally efficient alternatives to resource-intensive LLMs, particularly for users with limited resources. This section reviews advancements in smoothing, backoff, interpolation, and $n$-gram applications.

\vspace{1mm} \noindent\textbf{Smoothing Techniques.} One of the inherent limitations of $n$-gram models is their susceptibility to sparse data. In an $n$-gram language model, unseen sequences often receive a probability of zero, which can significantly affect the model's overall accuracy. Various smoothing techniques have been developed to address data sparsity in  models. One of the earliest methods, \textbf{Laplace Smoothing}, adds a small constant (typically 1) to each possible  count. This approach effectively prevents zero probabilities but may overestimate the likelihood of rare $n$-grams, reducing precision in some cases. A more advanced technique, \textbf{Kneser-Ney Smoothing}, improves upon this by assigning probabilities to unseen $n$-grams based on their occurrence across diverse contexts, making it particularly effective for rare word sequences. Additionally, \textbf{backoff and interpolation models} blend probabilities from different $n$-gram orders, providing a robust strategy to mitigate sparse data challenges. To address the limitations posed by sparse data, \textbf{backoff and interpolation models} have been widely adopted in $n$-gram modeling. \textbf{Backoff models} dynamically reduce the order of the $n$-gram when an observed sequence is not available, effectively ``backing off'' to lower-order $n$-grams. This allows the model to leverage smaller contexts when higher-order data is sparse. The simplicity of this approach ensures its wide applicability in computationally limited environments. In contrast to backoff models, \textbf{interpolation models}, blend probabilities from multiple $n$-gram lengths, assigning weights based on the reliability of the different orders. These methods mitigate the issue of sparse data by smoothing the probability estimates across different context sizes. Han et al.~\cite{han2024small} demonstrated that interpolation models can enhance language modeling tasks by incorporating multiple $n$-gram lengths effectively, while showing that fine-tuning smaller language models on task-specific data enables self-correction. Laplace smoothing has long been recognized as a foundational method~\cite{laplacesmoothingearlyref} while Kneser-Ney smoothing~\cite{kneser1995improved} and backoff-interpolation approaches~\cite{katz1987estimation} continue to enhance the accuracy and applicability of $n$-gram models.

\vspace{1mm} \noindent\textbf{Advanced $n$-gram Models in NLP.} Despite the recent surge in the development of neural language models such as transformers, $n$-gram models continue to be relevant in specific NLP tasks, particularly those requiring lower computational overhead and greater interpretability, like text generation, machine translation, and autocompletion. For applications such as autocompletion and basic \textbf{text generation}, $n$-gram models offer considerable speed and efficiency. Their lower computational cost makes them suitable for environments with limited processing power, where larger neural models like transformers may be overkill ~\cite{manning20introduction}. While neural models have set the state-of-the-art (SoTA) in \textbf{machine translation}, $n$-gram models still serve as a valuable alternative for small-scale tasks where computational transparency and speed are critical. For example, recent research by Zhao et al.~\cite{zhao2023survey} discuss recent advancements in large language models (LLMs) like GPT~\cite{achiam2023gpt} and LLaMA~\cite{touvron2023llama}, highlighting their superior accuracy in pre-training, fine-tuning, and evaluation, but also noting their high computational demands, whereas simpler models like $n$-grams remain useful for resource-constrained tasks due to their lower resource consumption.In terms of \textbf{resource efficiency}, Large Language Models (LLMs), such as GPT-3 and LLaMA-70B, require massive amounts of computational power for both training and inference. Zhou et al.~\cite{zhou2024survey} estimate that mainstream LLMs like LLaMA-70B, with 70 billion parameters, require around 140 GB of VRAM and multiple high-end GPUs, making them slow, resource-intensive, and inaccessible to many users, while simpler $n$-gram models offer a more efficient alternative in environments with limited hardware resources. In terms of \textbf{task-specific improvements}, recent studies suggest that even smaller $n$-gram models can be improved for specific tasks. Han et al.~\cite{han2024small} demonstrated that lightweight $n$-gram models can be fine-tuned on task-specific data to achieve higher accuracy. This technique balances the trade-offs between the computational efficiency of $n$-grams and the performance typically associated with larger LLMs. Future versions of HITgram could explore hybrid models combining $n$-grams with neural networks, potentially bridging performance gaps. Additionally, incorporating multilingual corpora and advanced tokenization strategies will further enhance its utility in diverse NLP applications.

\section{Motivation}
\label{sec:motivation}
Despite the dominance of LLMs, resource constraints highlight the need for platforms like HITgram to support lightweight, interpretable $n$-gram models. These models address challenges like data sparsity and resource efficiency using techniques like Laplace and Good-Turing smoothing.

\vspace{1mm}
\noindent\textbf{Long-Range Dependency Problems.} Traditional $n$-gram models are limited by their fixed-length context windows. For example, a trigram model only considers the two preceding words when predicting the next word. This fixed context prevents the model from capturing long-range dependencies, which are essential for understanding complex linguistic patterns. Models like transformers, by contrast, excel at capturing these dependencies but at a high computational cost [2]. HITgram seeks to provide a platform that allows experimentation with these limitations, encouraging the exploration of techniques that could extend the context window of $n$-gram models without compromising efficiency. $n$-gram models are limited by their fixed context window. For instance, a trigram model only considers the previous two words when predicting the next word. This short context window prevents the model from capturing long-range dependencies, which are crucial for understanding complex linguistic structures.

\vspace{1mm}
\noindent\textbf{Balancing Accuracy and Efficiency.} While $n$-gram models are computationally efficient, their predictive accuracy often lags behind more advanced models like LLMs. This discrepancy is particularly evident in complex tasks that require a nuanced understanding of linguistic structures. Enhancing the prediction accuracy of $n$-grams while maintaining their computational advantages is a key motivation for HITgram. Through features like configurable context lengths and built-in smoothing methods, HITgram aims to strike a balance between simplicity and performance, offering an effective solution for environments where computational resources are limited [3]. Despite their computational advantages, $n$-gram models often struggle with prediction accuracy compared to more sophisticated models like LLMs. There is a need to enhance $n$-gram models by incorporating advanced techniques like smoothing and context-sensitive weighting to improve their accuracy without sacrificing efficiency.

\vspace{1mm}
\noindent\textbf{The Need for Accessible, Interpretable Models.} Another driving factor behind HITgram is the increasing need for interpretable models~\cite{singh2023augmenting}. In applications where transparency and explainability are crucial e.g., legal, healthcare, or finance sectors -- $n$-gram models provide a clear advantage over opaque neural-based models. HITgram promotes the use of $n$-grams as interpretable, resource-efficient alternatives to LLMs, especially for those who require models to be both understandable and manageable without sacrificing too much predictive power.

\section{The HITGram Platform}
\label{sec:overview}
HITgram offers a user-friendly platform for $n$-gram model experimentation, supporting corpus management, customizable tokenization, and smoothing techniques. It provides a lightweight, efficient alternative to resource-intensive LLMs, ideal for resource-limited users. Overall, our HITgram platform allows users to: (a) upload text corpora in various acceptable file formats, (b) apply customizable tokenization strategies to split the text into tokens, (c) create $n$-gram models from the tokenized corpora, (d) utilize smoothing techniques to enhance model predictions, and (e) dynamically manage the text corpora by uploading new corpus for augmenting, removing, or replacing the existing ones.

\noindent Developed in Java~\cite{jana2005java} and utilizing Java Swing for its graphical user interface (GUI), HITgram delivers a robust environment for experimentation while remaining accessible. First, our platform allows for the creation of $n$-gram models from user-uploaded corpora in various file formats, like .txt, .pdf, etc. Users can easily upload and manage corpora through an interactive interface, enabling experimentation with different datasets and the integration of new data to refine models. HITgram's corpus management system offers flexibility, allowing datasets to be cleared or replaced to promote diverse experimental setups.

\vspace{1mm} \noindent\textbf{Built-in Tokenization Engine.} HITgram includes a built-in tokenizer that processes the uploaded corpus to break it into individual tokens. Based on the task that a user intends to perform such as \textit{sentence completion}, \textit{word completion}, or \textit{next-word prediction}, the choice of tokenization plays a crucial role. A user can apply customizable tokenization strategies that split text into words, sub-words (through stemming or lemmatization~\cite{boban2020sentence}), or characters. Once the $n$-gram model is constructed, users can input a sentence, and the platform predicts the next word(s) based on the trained model. For example, given the input ``\texttt{The cat}'', the model predicts the most likely next word based on the preceding tokens.
\textbf{Code snippet of Tokenization}
\begin{Verbatim}[baselinestretch=0.75,frame=single,fontsize=\footnotesize]
public List<String> tokenizeText(String text) {
  text = text.toLowerCase().replaceAll("[^a-zA-Z\\s]", ""); // Normalize
  return Arrays.asList(text.split("\\s+")); // Tokenize by space
}
\end{Verbatim}

\vspace{1mm} \noindent\textbf{Building and Customizing $\mathbf{n}$-gram Models.} The core of HITgram lies in building and customizing $n$-gram models. Users can build various types of $n$-gram models viz., unigrams, bigrams, trigrams, or higher-order $n$-grams -- each capturing different levels of contextual relationships in the text.

\noindent HITgram allows users to create different $n$-gram models based on the uploaded corpora. $n$-gram models work by generating sequences of $n$ tokens from the text. These models are categorized by the value of $n$: unigram (n = 1), bigram (n = 2), trigram (n = 3), and higher-order $n$-grams for larger $n$ values. The choice of $n$ affects the amount of context considered when predicting the next word or character in a sequence. For example, a trigram model (n = 3) would create two trigrams ``\texttt{The cat is}'' and ``\texttt{cat is sleeping}'' from the sentence ``\texttt{The cat is sleeping}''. From such $n$-length sequences, the context of the $n-1$ preceding words, along with the corresponding next word, is stored in a key-value format for efficient access and manipulation.

\noindent Overall, by building an $n$-gram model, we predict the likelihood of a word or sequence appearing in a particular context by analyzing the frequency of those sequences in the corpus. For example, in a unigram model, each word's probability is based solely on its frequency, while in a bigram model, the prediction depends on the preceding word. Trigram models extend this further by considering the previous two words. This probabilistic approach reveals dependencies within the text, enabling more accurate predictions.

\vspace{1mm} \noindent\textbf{Adaptive Context Lengths and Dynamic $n$-gram Model Updates.} HITgram introduces significant enhancements to traditional $n$-gram models, focusing on lightweight architecture and adaptive learning to support users with limited resources. Users can build $n$-gram models with various context lengths, from unigrams (n=1) to higher-order $n$-grams like bigrams (n=2) and trigrams (n=3). This flexibility allows users to balance context capture and computational efficiency. For instance, a trigram model captures more context than a bigram but requires more resources. We use the following lightweight Java code snippet (simplified) to dynamically update an existing $n$-gram model:

{\begin{Verbatim}[baselinestretch=0.75,frame=single,fontsize=\footnotesize]
// Tokenize the new corpus
String[] newWords = newCorpus.tokenize(); 
// Update existing $n$-gram model 
for (int i = 0; i <= newWords.length - n; i++) { 
  StringBuilder nGram = new StringBuilder(); 
  for (int j = 0; j < n - 1; j++) { 
    nGram.append(newWords[i + j]).append(" "); 
  } 
  String nextWord = newWords[i + n - 1];
  // Use putIfAbsent for initialization and merge for counting
  nGramModel.putIfAbsent(nGram.toString().trim(), new HashMap<>());
  nGramModel.get(nGram.toString().trim()).
    merge(nextWord, 1, Integer::sum); 
} 
\end{Verbatim}
}

\noindent Additionally, an intuitive GUI slider allows users to modify the value of $n$ easily, providing interactive insights into how context length adjustments affect predictions.

\vspace{1mm} \noindent\textbf{Enhancing Rare Sequence Completion with Smoothing.} One of the key challenges in $n$-gram models is the presence of unseen word pairs or triplets, which can lead to zero probabilities for these rare sequences and significantly degrade the model's performance. For example, consider a corpus consisting of sentences like ``\texttt{I love natural language processing models immensely as it is fun}''. If our model generates bigrams and we encounter a next-word prediction task for the phrase ``\texttt{natural processing was}'', we encounter an unseen pair, leading to zero probabilities.

\noindent To address this challenge of unseen words in $n$-gram models, HITgram employs smoothing techniques. Smoothing enhances the model's ability to generalize to previously unseen data by ensuring that every possible $n$-gram maintains a non-zero probability. It adjusts the probability distribution, ensuring all word combinations, including those not present in the training data, receive a non-zero probability. This capability is crucial for real-world applications where not all word combinations are available in the training dataset. Our HITgram platform offers various smoothing options, such as Laplace, Add-k, and Good-Turing smoothing, enabling the $n$-gram model to effectively manage scenarios involving rare sequences, thereby enhancing its predictive power and adaptability.

\begin{itemize}
    \item \textbf{Additive (Laplace) Smoothing} involves adding a small constant (typically 1) to the count of each $n$-gram, including unseen ones, thereby preventing any probability from being zero. For instance, in a bigram model, the probability of a word given its predecessor is computed as $P(w_i|w_{i-1}) = \frac{\text{Count}(w_i,w_{i-1}) + 1}{\text{Count}(w_{i-1}) + V}$ where $\text{Count}(w_{i-1})$ is the count of the first word in the bigram and $V$ is the total number of unique words in the vocabulary. From the example above, although the bigram (\texttt{was}, \texttt{fun}) has no corpus occurrences, Laplace smoothing guarantees that it will still yield a non-zero probability. 
    \item \textbf{Add-k Smoothing} is a generalized version of Laplace smoothing where a smaller constant k (e.g., 0.01) is added instead of 1. This approach allows for finer control over the smoothing applied. The formula for Add-k Smoothing is: $P(w_i|w_{i-1}) = \frac{\text{Count}(w_i,w_{i-1})+k}{\text{Count}(w_{i-1})+kV}$. By using a smaller value for k,  the method assigns a more realistic, albeit reduced, probability to unseen $n$-grams. 
    \item \textbf{Good-Turing Smoothing} is a more sophisticated technique that estimates the probability of unseen $n$-grams based on the frequency of $n$-grams that have been observed once, twice, etc. The formula is: $P(w_i|w_{i-1}) = \frac{N_1}{N}$, where $N_1$ is the count of bigrams that appear exactly once, and $N$ is the total number of bigrams. This method is particularly effective in large corpora, as it redistributes probability more efficiently among unseen $n$-grams. 
\end{itemize}

\vspace{1mm} \noindent\textbf{Context-Sensitive Weighting.}
To address the challenge of unseen word sequences that may yield zero probabilities in standard $n$-gram models, HITgram employs smoothing techniques like Laplace smoothing. This enables the model to assign non-zero probabilities to $n$-grams absent from the training data. Furthermore, HITgram introduces context-sensitive weighting, which enhances the model’s resilience against sparse or incomplete data, boosting prediction accuracy. To stabilize weights, particularly for infrequent $n$-grams, a logarithmic transformation is applied. A simplified code snippet for this is shown below: 
{\begin{Verbatim}[baselinestretch=0.75,frame=single,fontsize=\footnotesize]
private static double calculateWeight(String nGramKey) {
  double frequency = tokenFrequencyMap.getOrDefault(nGramKey, 0);
  // Logarithmic transformation to stabilize weights
  return Math.log(1 + frequency); // Smoothing the frequency
}
\end{Verbatim}
}

\noindent Thus, instead of treating all $n$-grams uniformly, the platform enables users to assign different weights to word sequences based on their frequency or significance within the training corpus. This enhances the model's ability to predict rare yet essential word combinations.

\vspace{1mm} \noindent\textbf{Dynamic Corpus Management for Adaptive Text Completion.}
Users can upload multiple corpora for concatenation or replacement, enabling incremental learning and diverse text data experiments.

\vspace{1mm} \noindent\textbf{Algorithmic Improvements for Computational Efficiency.} We aim to improve the efficiency of generating and predicting $n$-gram models, especially for users with limited computing resources. The HITgram platform achieves this through various algorithmic optimizations, as outlined below.

\noindent HITgram uses key-value structures and pre-computed probabilities to optimize memory and retrieval, supporting real-time tasks like sentence completion. Pruning methods and efficient tokenization algorithms enhance computational efficiency for resource-limited environments.

\noindent A simplified code snippet demonstrating this is provided below:

{\begin{Verbatim}[baselinestretch=0.75,frame=single,fontsize=\footnotesize]
private static void pruneLowFrequencyNGrams(int threshold) {
  nGramModel.entrySet().removeIf(entry -> entry.getValue().values()
       .stream()
       .mapToInt(Integer::intValue)
       .sum() < threshold);
}
\end{Verbatim}
}

\noindent Future HITgram versions could integrate neural embeddings to capture short and long-range dependencies, multilingual corpora, and advanced tokenization. Features like model saving and cross-entropy metrics will further bridge the gap between lightweight and complex models.

\section{Experimental Results}
\label{sec:expt}

In this section, we present the results of experiments conducted on the HITgram platform, which focuses on lightweight $n$-gram language models. The system allows users to upload a corpus, build an $n$-gram model, and predict the next word sequence based on input tokens. Our experiments evaluate performance across various corpus sizes and assess the system’s effectiveness in tasks such as tokenization, model construction, next-word prediction, and sentence completion. The results highlight the practicality of $n$-gram models for users with limited computational resources compared to large-scale language models (LLMs).
HITgram tokenizes English text at up to 50,000 tokens/second and generates 2-grams from a 320 MB corpus in under 63 seconds, showcasing its efficiency in resource-constrained environments. We used \textbf{perplexity} as primary evaluation metric. Perplexity, a key performance metric for $n$-gram models, evaluates prediction effectiveness on validation text: 
\[ \text{PP(W)} = \sqrt[N]{{P(w_1, w_2, \ldots, w_N)}} = \sqrt[N]{\prod_{i=1}^{N} \frac{1}{P(w_i \mid w_1, w_2, \ldots, w_{i-1})}} \] where, $\text{PP(W)}$ represents the perplexity of the test set W, where a lower perplexity signifies a more effective predictive model. ${P(w_1, w_2, \ldots, w_N)}$ is the joint probability of the complete sequence of words $w_1, w_2, \ldots, w_N$ in the test set, N is the total count of words in the test dataset and ${P(w_i \mid w_1, w_2, \ldots, w_{i-1})}$ is the conditional probability of ${w_i}$ based on the sequence of immediately preceding words ${w_1, w_2, \ldots, w_{i-1}}$. The analysis is based on the sentence: "this is a". Table \ref{tab:perplexityexpt} shows the file size (in MB), bigram perplexity, and trigram perplexity for different file sizes.

\setlength{\tabcolsep}{8pt}
\begin{table}[t]
\centering
\caption{Perplexity values of test set of words for language models}
\label{tab:perplexityexpt}
\vspace{1mm}
\resizebox{0.4\textwidth}{!}{%
\begin{tabular}{c|rr}
\hline
Corpus Size (MB) & \multicolumn{2}{c}{Perplexity **} \\
\cline{2-3}
 & bigram & trigram \\
\hline
997 & 4.4821 & 2.3886 \\
900 & 4.0598 & 2.2009 \\
800 & 4.2325 & 2.1666 \\
720 & 4.1310 & 2.0156 \\
660 & 4.3243 & 2.2325 \\
450 & 4.2337 & 2.2440 \\
400 & 4.1536 & 2.3896 \\
320 & 4.2054 & 2.5989 \\
\hline
\end{tabular}
}\endnote\tiny{** results may vary depending on machine configuration}
\end{table}

\setlength{\tabcolsep}{8pt}
\begin{table}[t]
\centering
\caption{\textbf{Performance Analysis}: Time to build $n$-gram models}
\label{tab:ngram_times}
\vspace{1mm}
\resizebox{0.7\textwidth}{!}{%
\begin{tabular}{r|r|rrrr}
\hline
Corpus Size (KB) & Time to Load \& Tokenize (ms)** & \multicolumn{4}{c}{Time to Build $n$-gram Model (ms)**} \\
\cline{3-6}
 & & $\,n=1\,$ & $\,n=2\,$ & $\,n=3\,$ & $n=4$ \\
\hline
4898.76 & 473.5 & 55.1 & 19.5 & 16.1 & 15.3 \\
5800.846 & 648.3 & 16.5 & 32.6 & 16.1 & 24.4 \\
6526.48 & 899.2 & 15.5 & 32.6 & 34.4 & 31.5 \\
7245.54 & 1695 & 24.5 & 31.7 & 32.7 & 30.8 \\
10302.58 & 2689 & 25.3 & 32.2 & 31.7 & 34.5 \\
11922.73 & 2779.3 & 75.2 & 77.4 & 109.7 & 121.5 \\
13051.37 & 2918.7 & 173.6 & 286.5 & 449.7 & 452.3 \\
13273.6 & 4001.5 & 236.7 & 231.5 & 484.3 & 450.6 \\
\hline
\end{tabular}
}\endnote\tiny{** measured using AMD Ryzen mid-range performance processors with 4 cores, 8GB RAM.}
\end{table}

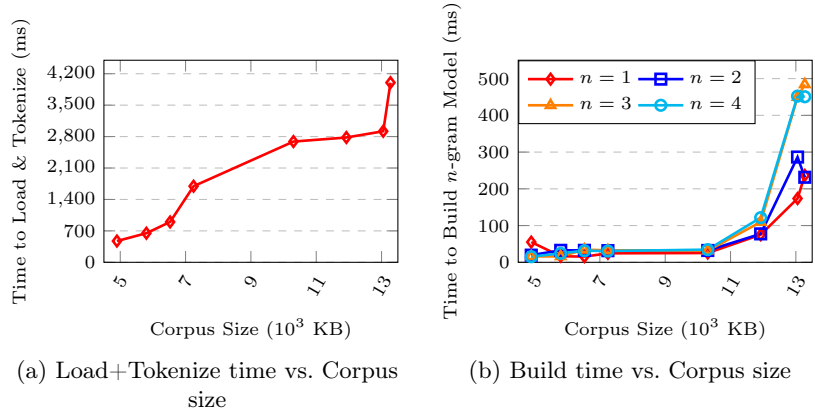
\begin{figure}[h!]%
\vspace{-7mm}
    \centering
    \subfloat[\centering Load+Tokenize time vs. Corpus size]{\pgfplotsset{width=0.45\textwidth,height=0.35\textwidth,compat=1.9}\begin{tikzpicture}[font=\scriptsize]
\begin{axis}[
scaled ticks=false,
xlabel={Corpus Size ($10^3$ KB)},
ylabel={Time to Load \& Tokenize (ms)},
xmin=4500, xmax=13500,
ymin=0, ymax=4500,
scaled x ticks=base 10:-3,
xtick scale label code/.code={},
xtick={5000,7000,9000,11000,13000},
ytick={0,700,1400,2100,2800,3500,4200},
xticklabel style={rotate=60},
ymajorgrids=true,
grid style=dashed
]

\addplot[
color=red,
mark=diamond,
mark size=2pt,
line width=1pt
]
coordinates {
(4898.76,473.5)
(5800.846,648.3)
(6526.48,899.2)
(7245.54,1695)
(10302.58,2689)
(11922.73,2779.3)
(13051.37,2918.7)
(13273.6,4001.5)
};

\end{axis}
\end{tikzpicture}}%
    \hspace{2mm}
    \subfloat[\centering Build time vs. Corpus size]{\pgfplotsset{width=0.45\textwidth,height=0.35\textwidth,compat=1.9}\begin{tikzpicture}[font=\scriptsize]
\begin{axis}[
xlabel={Corpus Size ($10^3$ KB)},
ylabel={Time to Build $n$-gram Model (ms)},
xmin=4500, xmax=13500,
ymin=0, ymax=550,
scaled x ticks=base 10:-3,
xtick scale label code/.code={},
xtick={5000,7000,9000,11000,13000},
ytick={0,100,200,300,400,500},
xticklabel style={rotate=60},
legend columns=2,
legend style={at={(0.0,1.0)}, anchor=north west,font=\scriptsize},
ymajorgrids=true,
grid style=dashed
]

\addlegendentry{$n=1$}
\addplot[
color=red,
mark=diamond,
mark size=2pt,
line width=1pt
]
coordinates {
(4898.76,55.1)
(5800.846,16.5)
(6526.48,15.5)
(7245.54,24.5)
(10302.58,25.3)
(11922.73,75.2)
(13051.37,173.6)
(13273.6,236.7)
};

\addlegendentry{$n=2$}
\addplot[
color=blue,
mark=square,
mark size=2pt,
line width=1pt
]
coordinates {
(4898.76,19.5)
(5800.846,32.6)
(6526.48,32.6)
(7245.54,31.7)
(10302.58,32.2)
(11922.73,77.4)
(13051.37,286.5)
(13273.6,231.5)
};

\addlegendentry{$n=3$}
\addplot[
color=orange,
mark=triangle,
mark size=2pt,
line width=1pt
]
coordinates {
(4898.76,16.1)
(5800.846,16.1)
(6526.48,34.4)
(7245.54,32.7)
(10302.58,31.7)
(11922.73,109.7)
(13051.37,449.7)
(13273.6,484.3)
};

\addlegendentry{$n=4$}
\addplot[
color=ProcessBlue,
mark=o,
mark size=2pt,
line width=1pt
]
coordinates {
(4898.76,15.3)
(5800.846,24.4)
(6526.48,31.5)
(7245.54,30.8)
(10302.58,34.5)
(11922.73,121.5)
(13051.37,452.3)
(13273.6,450.6)
};

\end{axis}
\end{tikzpicture}}%
    \caption{\textbf{Plot of the time taken by HITgram:} Time taken to load and build $n$-gram models, by varying corpus size and $n$.}
    \label{fig:perfexample}%
\vspace{-5mm}
\end{figure}

\vspace{1mm}
\noindent\textbf{Dataset Description.}
Experiments used diverse novel-based corpora from \url{https://www.kaggle.com/}  and \url{http://textfiles.com/etext/FICTION/}, including \textit{anna\_karenina.txt}, \textit{warpeace.txt}, \textit{quixote.txt}, and \textit{lesms10.txt}.

\vspace{1mm} \noindent\textbf{Experimental Setup.}
HITgram constructs $n$-gram models with varying $n$ values, efficiently processing PDFs (e.g., a 0.8 GB file to 10,302 KB in 2,689 ms).

\vspace{1mm} \noindent\textbf{Corpus File and Tokenization.} After uploading a corpus (PDF or text), HITgram tokenizes the text into individual tokens for $n$-gram model construction. The time for tokenization depends on the corpus size; for example, tokenizing a 13MB text file took 4001 ms, demonstrating efficiency in handling large datasets.

\vspace{1mm} \noindent\textbf{$n$-gram Model Construction.}
After tokenization, the $n$-gram model is constructed based on the chosen value of $n$. The model utilizes Laplace Smoothing to handle unseen word combinations, ensuring a more accurate prediction for less frequent sequences. Table ~\ref{tab:ngram_times} shows that $n$-gram model construction time increases linearly with corpus size and $n$, demonstrating HITgram’s scalability. For smaller corpora, the $n$-gram model can be built in a relatively short time, even for larger $n$ values. However, for larger corpora, the time increases substantially, particularly for $n$ values of 3 and 4. Comparative graphs between corpus size and the time to load and tokenize, as well as the time to build the $n$-gram model for varying $n$ values, are presented in Figure~\ref{fig:perfexample}. Figure~\ref{fig:perfexample} indicates the scaling behavior of HITgram with respect to corpus size and n values. The results highlight HITgram's advantage in efficiently handling medium-sized corpora while maintaining performance with larger ones. Table ~\ref{tab:ngram_seq} shows an example of the $n$-gram sequences built from the corpus.

\setlength{\tabcolsep}{8pt}
\begin{table}[t]
\centering
\caption{\textbf{Example of $n$-gram modeling}: Frequency of corpus sequences}
\label{tab:ngram_seq}
\vspace{1mm}
\begin{tabular}{c|c|r}
\hline
\textbf{$\mathbf{n-1}$ words (Key)} & \textbf{Next word (Value)} & \textbf{Frequency} \\
\hline
Artificial Intelligence & is & 52 \\
is transforming & industries & 28 \\
the future & of & 43 \\
\hline
\end{tabular}
\end{table}

\begin{figure}[t]
    \centering
    \includegraphics[width=0.65\textwidth]{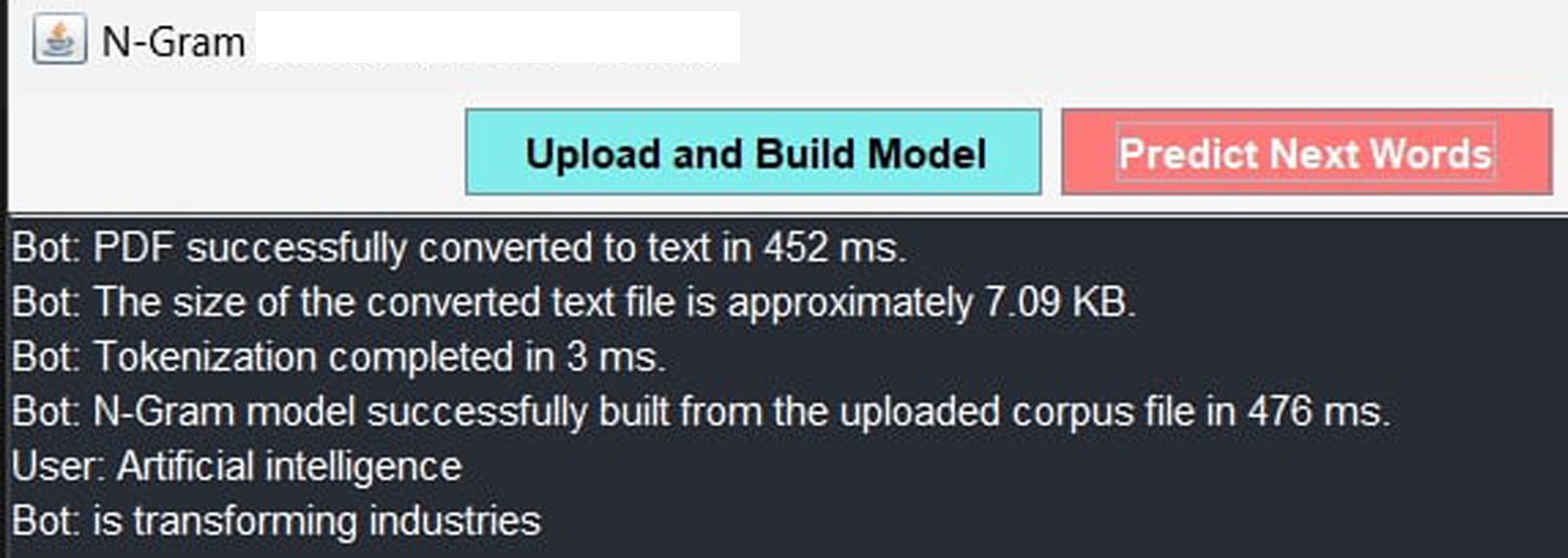}
    \caption{\textbf{HITgram in Action}: A text corpus generated from a user-uploaded PDF, followed by downstream processing.}
    \label{fig:screenshot}
\vspace{-5mm}
\end{figure}


\vspace{1mm} \noindent\textbf{Prediction of Next Words. }
To evaluate the platform's prediction capability, we tested it by inputting the phrase ``\texttt{Artificial Intelligence}'' and requesting five subsequent tokens. Based on the trigram model, the system predicted:
$\text{\texttt{Artificial Intelligence is transforming industries worldwide}.}
$

\noindent A screenshot of the activities has been captured in Figure~\ref{fig:screenshot}. This prediction leverages the highest probability derived from the $n$-gram model, utilizing Laplace smoothing to avoid zero probabilities for unseen sequences. The probability for each predicted word is calculated as follows:

$P(w | w_{n-1}) = \frac{\text{Frequency of } w + 1}{\text{Total Occurrences of } (n-1) \text{ Words} + \text{Vocabulary Size}}$

\noindent This method ensures non-zero probabilities for sequences not present in the corpus, allowing for accurate predictions that reflect common continuations found in the training data. After constructing the $n$-gram model, users can specify how many words to predict based on the $n$-gram probability distribution. 


\vspace{1mm} \noindent\textbf{Reproducibility.}
For those interested in exploring the source code or contributing to the project, the code repository is available on GitHub at: \url{https://github.com/chandan789maity/HITgram}. This repository also contains an installer that helps to setup HITgram on Windows$^\text{TM}$ system.

\section{Conclusion and Future Work}
\label{sec:concl}
We demonstrate that HITgram has the potential of $n$-gram language modeling as a lightweight alternative to large language models (LLMs), particularly for users with limited computing resources. By providing tools that include context-sensitive weighting, Laplace smoothing, and flexible corpus management, the platform allows for efficient and accurate text modeling. With its adaptability and user-friendly interface, HITgram makes $n$-gram models approachable and valuable for those in resource-constrained environments. Its success in handling unseen word sequences and incremental learning underscores its significance as a powerful tool for foundational natural language processing (NLP) tasks.

\noindent Looking ahead, we believe that HITgram has great potential. Advanced smoothing, multilingual corpora, and parallel processing can enhance efficiency. Although HITgram is optimized for English text, future versions will incorporate multilingual corpora and context-aware tokenization for more diverse applications. Additionally, parallel processing for large datasets will enhance scalability. Dynamic $n$-gram models and collaboration features will broaden its use. Integrating advanced NLP models will bring richer predictions. Future updates may include flexible tokenization, performance metrics like cross entropy, and model-saving options. These improvements will make HITgram even more valuable, bridging the gap between lightweight and complex language models.

\bibliographystyle{splncs04}
\bibliography{mybib}
\end{document}